\documentclass[11pt,a4paper]{article}
\usepackage[hyperref]{acl2019}
\usepackage{times}
\usepackage{latexsym}
\usepackage{booktabs}
\usepackage{graphicx}
\usepackage{amsmath}
\usepackage{float}
\usepackage{graphics}

\usepackage{url}

\aclfinalcopy % Uncomment this line for the final submission
\def\aclpaperid{66} %  Enter the acl Paper ID here

\newcommand{\architecturename}{HERMIT}
\newcommand{\datasetname}{ROMULUS}

\title{Hierarchical Multi-Task Natural Language Understanding for Cross-domain Conversational AI: HERMIT NLU}

\author{Andrea Vanzo \\
Interaction Lab\\
Heriot-Watt University\\
\texttt{a.vanzo@hw.ac.uk} \\\And
Emanuele Bastianelli \\
Interaction Lab\\
Heriot-Watt University\\
\texttt{e.bastianelli@hw.ac.uk} \\\And
Oliver Lemon \\
Interaction Lab\\
Heriot-Watt University\\
\texttt{o.lemon@hw.ac.uk} \\
}

\date{}

\begin{document}
\maketitle
\begin{abstract}
	We present a new neural architecture for wide-coverage Natural Language Understanding in Spoken Dialogue Systems. We develop a hierarchical multi-task architecture, which delivers a multi-layer representation of sentence meaning (i.e., Dialogue Acts and Frame-like structures). The architecture is a hierarchy of self-attention mechanisms and BiLSTM encoders followed by CRF tagging layers.
	We describe a variety of experiments, showing that our approach obtains promising results on a dataset annotated with Dialogue Acts and Frame Semantics. Moreover, we demonstrate its applicability to a different, publicly available NLU dataset annotated with domain-specific intents and corresponding semantic roles, providing overall performance higher than state-of-the-art tools such as RASA, Dialogflow, LUIS, and Watson.
	For example, we show an average $4.45\%$ improvement in entity tagging F-score over Rasa, Dialogflow and LUIS.
	
\end{abstract}

\section{Introduction}
\label{sec:intro}

Research in Conversational AI (also known as Spoken Dialogue Systems) has  applications ranging from home devices to robotics, and has a growing presence in industry.
A key problem in real-world Dialogue Systems is Natural Language Understanding (NLU) -- the process of extracting structured representations of meaning from user utterances. In fact, the effective extraction of semantics is an essential feature, being the entry point of any Natural Language interaction system.
Apart from challenges given by the inherent complexity and ambiguity of human language, other challenges arise whenever the NLU has to operate over multiple domains.
In fact, interaction patterns, domain, and language vary depending on the device the user is interacting with. For example, chit-chatting and instruction-giving for executing an action are different processes in terms of language, domain, syntax and interaction schemes involved. And what if the user combines  two interaction domains: ``\textit{play some music, but first what's the weather tomorrow}''?

In this work, we present \architecturename{}, a HiERarchical MultI-Task Natural Language Understanding architecture\footnote{\url{https://gitlab.com/hwu-ilab/hermit-nlu}}, designed for effective semantic parsing of domain-independent user utterances, extracting meaning representations in terms of high-level intents and frame-like semantic structures.
With respect to previous approaches to NLU for SDS, HERMIT stands out for being a cross-domain, multi-task architecture, capable of recognising multiple intents/frames in an utterance. HERMIT also shows better performance with respect to current state-of-the-art commercial systems. Such a novel combination of requirements is discussed below.

\paragraph{Cross-domain NLU}
A cross-domain dialogue agent must be able to handle heterogeneous types of conversation, such as chit-chatting, giving directions, entertaining, and triggering domain/task actions. A domain-independent and rich meaning representation is thus required to properly capture the intent of the user. Meaning is modelled here through three layers of knowledge: dialogue acts, frames, and frame arguments. Frames and arguments can be in turn mapped to domain-dependent intents and slots, or to Frame Semantics' \cite{fillmore1976frame} structures (i.e. semantic frames and frame elements, respectively), which allow handling of heterogeneous domains and language.

\paragraph{Multi-task NLU}
Deriving such a multi-layered meaning representation can be approached through a multi-task learning approach. Multi-task learning has found success in several NLP problems \cite{hashimoto2017, Strubel18:emnlp}, especially with the recent rise of Deep Learning. Thanks to the possibility of building complex networks, handling more tasks at once has been proven to be a successful solution, provided that some degree of dependence holds between the tasks. Moreover, multi-task learning allows the use of  different datasets to train sub-parts of the network \cite{sanh2018hmtl}. Following the same trend, \architecturename{} is a hierarchical multi-task neural architecture which is able to deal with the three tasks of tagging dialogue acts, frame-like structures, and their arguments in  parallel.
The network, based on self-attention mechanisms, seq2seq bi-directional Long-Short Term Memory (BiLSTM) encoders, and CRF tagging layers, is hierarchical in the sense that information output from earlier layers flows through the network, feeding following layers to solve downstream dependent tasks.

\paragraph{Multi-dialogue act and -intent NLU}
Another degree of complexity in NLU is represented by the granularity of knowledge that can be extracted from an utterance. Utterance semantics is often rich and expressive: approximating meaning to a single user intent is often not enough to convey the required information. As opposed to the traditional single-dialogue act and single-intent view in previous work \cite{Guo14:SLTW, LiuL16:interspeech, hakkani-tur2016multi}, \architecturename{} operates on a meaning representation that is multi-dialogue act and multi-intent. In fact, it is possible to model an utterance's meaning through multiple dialogue acts and intents at the same time. For example, the user would be able both to request tomorrow's weather and listen to his/her favourite music with just a single utterance.\\

\noindent
A further requirement is that for practical application the system should be \textbf{competitive with state-of-the-art}: we evaluate \architecturename's effectiveness by running several empirical investigations. We perform a robust test on a publicly available NLU-Benchmark (NLU-BM) \cite{Liu2019} containing 25K cross-domain utterances with a conversational agent. The results obtained show a performance higher than well-known off-the-shelf tools (i.e., Rasa, DialogueFlow, LUIS, and Watson).  
The contribution of the different network components is then highlighted through an ablation study. We also test \architecturename{} on the smaller Robotics-Oriented MUltitask Language UnderStanding (\datasetname) corpus, annotated with Dialogue Acts and Frame Semantics.
\architecturename{} produces promising results for the application in a real scenario.

\section{Related Work}
Much  research on Natural (or Spoken, depending on the input) Language Understanding has been carried out in the area of Spoken Dialogue Systems \cite{chen17:SIGKDD}, where the advent of statistical learning has led to the application of many data-driven approaches \cite{lemon2012:book}. In recent years, the rise of deep learning models has further improved the state-of-the-art. Recurrent Neural Networks (RNNs) have proven to be particularly successful, especially uni- and bi-directional LSTMs and Gated Recurrent Units (GRUs). The use of such deep architectures has also fostered the development of joint classification models of intents and slots.
Bi-directional GRUs are applied in \cite{Zhang16:ijcai}, where the hidden state of each time step is used for slot tagging in a seq2seq fashion, while the final state of the GRU is used for intent classification. The application of attention mechanisms in a BiLSTM architecture is investigated in \cite{LiuL16:interspeech}, while the work of \cite{Chen16:interspeech} explores the use of memory networks \cite{Sukhbaatar15:nips} to exploit encoding of historical user utterances to improve the slot-filling task. Seq2seq with self-attention is applied in \cite{Li2018:emnlp}, where the classified intent is also used to guide a special gated unit that contributes to the slot classification of each token.

One of the first attempts to jointly detect domains in addition to intent-slot tagging is the work of \cite{Guo14:SLTW}. An utterance syntax is encoded through a Recursive NN, and it is used to predict the joined domain-intent classes. Syntactic features extracted from the same network are used in the per-word slot classifier. The work of \cite{hakkani-tur2016multi} applies the same idea of \cite{Zhang16:ijcai}, this time using a context-augmented BiLSTM, and performing domain-intent classification as a single joint task. As in \cite{Chen16:interspeech}, the history of user utterances is also considered in \cite{Bapna17:sigdial}, in combination with a dialogue context encoder. A two-layer hierarchical structure made of a combination of BiLSTM and BiGRU is used for joint classification of domains and intents, together with slot tagging. \cite {Rastogi18:acl} apply multi-task learning to the dialogue domain. Dialogue state tracking, dialogue act and intent classification, and slot tagging are jointly learned. Dialogue states and user utterances are encoded to provide hidden representations, which jointly affect all the other tasks.

Many previous systems are trained and compared over the ATIS (Airline Travel Information Systems) dataset \cite{Price:1990:ESL:116580.116612}, which covers only the flight-booking domain. Some of them also use bigger, not publicly available datasets, which appear to be similar to the NLU-BM in terms of number of intents and slots, but they cover no more than three or four domains. Our work stands out for its more challenging NLU setting, since we are dealing with a higher number of domains/scenarios (18), intents (64) and slots (54) in the NLU-BM dataset, and dialogue acts (11), frames (58) and frame elements (84) in the \datasetname{} dataset. Moreover, we propose a multi-task hierarchical architecture, where each layer is trained to solve one of the three tasks. Each of these is tackled with a seq2seq classification using a CRF output layer, as in \cite{sanh2018hmtl}.

The NLU problem has been studied also on the Interactive Robotics front, mostly to support basic dialogue systems, with few dialogue states and tailored for specific tasks, such as semantic mapping \cite{Kruijff:07www}, navigation \cite{kollar10:hri,bothe18:icsr}, or grounded language learning \cite{chai2016collaborative}. However, the designed approaches, either based on formal languages or data-driven, have never been shown to scale to real world scenarios. The work of \cite{hatori18:icra} makes a step forward in this direction. Their model still deals with the single `pick and place' domain, covering no more than two intents, but it is trained on several thousands of examples, making it able to manage more unstructured language. An attempt to manage a higher number of intents, as well as more variable language, is represented by the work of \cite{bastianelli:16ijcai} where the sole Frame Semantics is applied to represent user intents, with no Dialogue Acts.

\section{Jointly parsing dialogue acts and frame-like structures}
\label{sec:architecture}
The identification of Dialogue Acts (henceforth DAs) is required to drive the dialogue manager to the next dialogue state. General frame structures (FRs) provide a reference framework to capture user intents, in terms of required or desired actions that a conversational agent has to perform. Depending on the level of abstraction required by an application, these can be interpreted as more domain-dependent paradigms like \textit{intent}, or to shallower representations, such as \textit{semantic frames}, as conceived in FrameNet \cite{Baker:98framenet}. From this perspective, semantic frames represent a versatile abstraction that can be mapped over an agent's capabilities, allowing also the system to be easily extended with new functionalities without requiring the definition of new ad-hoc structures. Similarly, frame arguments (ARs) act as \textit{slots} in a traditional intent-slots scheme, or to \textit{frame elements} for semantic frames.

\begin{table*}[ht]
	\centering
	\small
	\begin{tabular}{lcccccc}
		& \textit{Where} & \textit{can} & \textit{I} & \textit{find} & \textit{Starbucks} & \textit{?} \\  
		DAs & \textsc{B-Req\_info} & \textsc{I-Req\_info} & \textsc{I-Req\_info} & \textsc{I-Req\_info} & \textsc{I-Req\_info} & \textsc{O} \\
		FRs & \textsc{B-}\textit{Locating} & \textsc{I-}\textit{Locating} & \textsc{I-}\textit{Locating} & \textsc{I-}\textit{Locating} & \textsc{I-}\textit{Locating} & \textsc{O} \\
		ARs & \textsc{O} & \textsc{O} & \textsc{B-Cognizer} & \textsc{B-Lexical\_unit} & \textsc{B-Entity} & \textsc{O} \\ 
	\end{tabular}
	\captionof{figure}{Dialogue Acts (DAs), Frames (FRs -- here semantic frames) and Arguments  (ARs -- here frame elements) IOB2 tagging for the sentence \textit{Where can I find Starbucks?}} \label{fig:tagging}
\end{table*}

In our work, the whole process of extracting a complete semantic interpretation as required by the system is tackled with a multi-task learning approach across DAs, FRs, and ARs. Each of these tasks is modelled as a seq2seq problem, where a task-specific label is assigned to each token of the sentence according to the IOB2 notation \cite{Sang:1999:RTC:977035.977059}, with ``B-'' marking the Beginning of the chunk, ``I-'' the tokens Inside the chunk while ``O-'' is assigned to any token that does not belong to any chunk. Task labels are drawn from the set of classes defined for DAs, FRs, and ARs. 
Figure \ref{fig:tagging} shows an example of the tagging layers over the sentence \textit{Where can I find Starbucks?}, where Frame Semantics has been selected as underlying reference theory.

\subsection{Architecture description}
The central motivation behind the proposed architecture is that there is a dependence among the three tasks of identifying DAs, FRs, and ARs. 
The relationship between tagging frame and arguments appears more evident, as also developed in theories like Frame Semantics -- although it is defined independently by each theory.
However, some degree of dependence also holds between the DAs and FRs. For example, the FrameNet semantic frame \textit{Desiring}, expressing a desire of the user for an event to occur, is more likely to be used in the context of an \textsc{Inform} DA, which indicates the state of notifying the agent with an information, other than in an \textsc{Instruction}. This is clearly visible in interactions like ``\textit{I'd like a cup of hot chocolate}'' or ``\textit{I'd like to find a shoe shop}'', where the user is actually notifying the agent about a desire of hers/his.

\begin{figure}[t]
	\centering
	\includegraphics[width=1.0\linewidth]{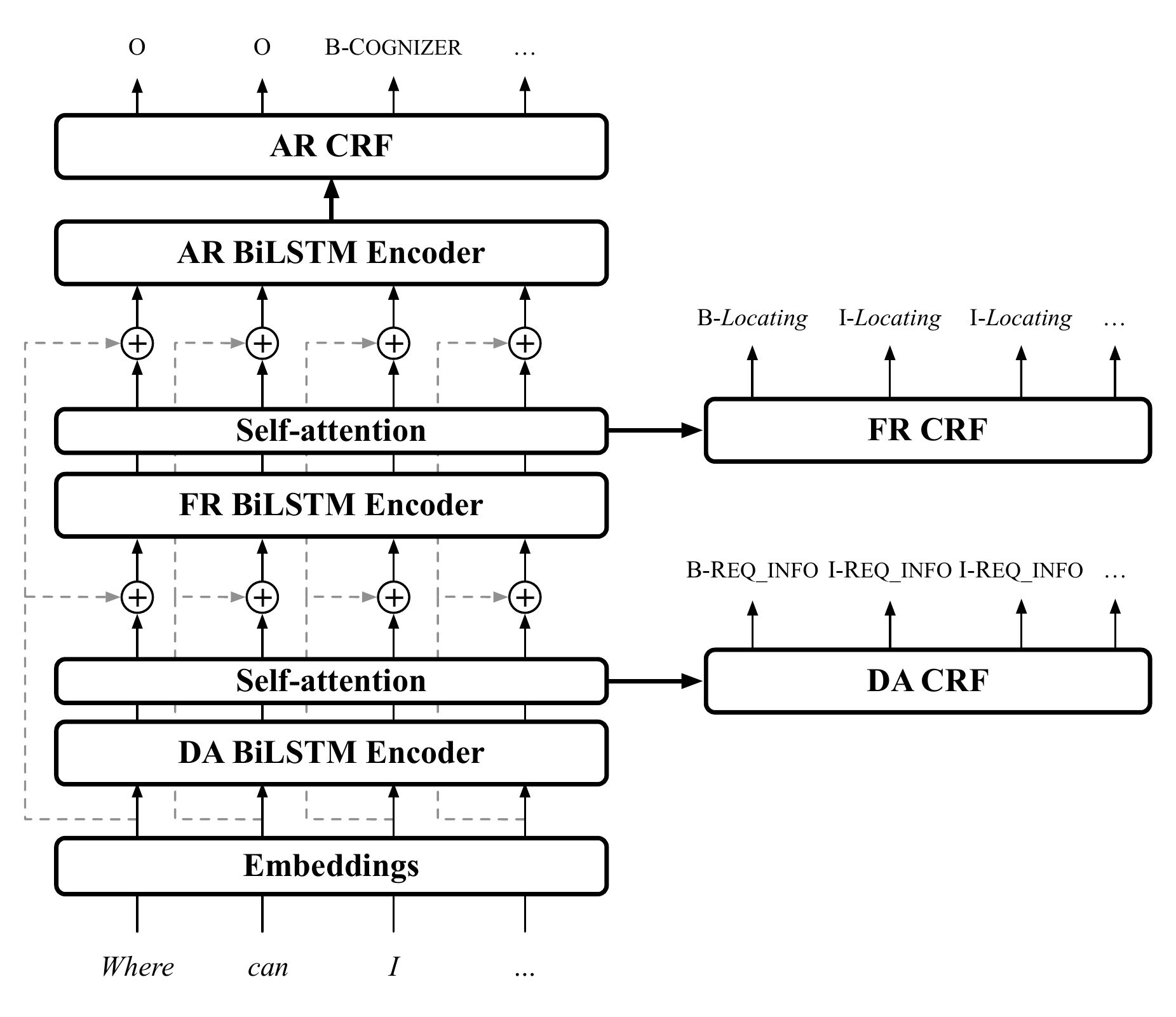}
	\caption{HERMIT Network topology}
	\label{fig:network}
\end{figure}

In order to reflect such inter-task dependence, the classification process is tackled here through a hierarchical multi-task learning approach. We designed a multi-layer neural network, whose architecture is shown in Figure \ref{fig:network}, where each layer is trained to solve one of the three tasks, namely labelling dialogue acts ($DA$ layer), semantic frames ($FR$ layer), and frame elements ($AR$ layer). The layers are arranged in a hierarchical structure that allows the information produced by earlier layers to be fed to downstream tasks.

The network is mainly composed of three BiLSTM \cite{Schuster97:transigproc} encoding layers. A sequence of input words is initially converted into an embedded representation through an ELMo embeddings layer \cite{peters2018}, and is fed to the $DA$ layer. The embedded representation is also passed over through shortcut connections \cite{hashimoto2017}, and concatenated with both the outputs of the $DA$ and $FR$ layers. Self-attention layers \cite{Zheng:2018:OOA:3219819.3219839} are placed after the $DA$ and $FR$ BiLSTM encoders. Where $w_t$ is the input word at time step $t$ of the sentence $\textbf{\textrm{w}} = (w_1, ..., w_T)$, the architecture can be formalised by:
\begin{gather*}
	\small
	e_t = ELMo(w_t), \; s_t^{DA} = BiLSTM(e_t)\\
	a_t^{DA} = SelfAtt(s_t^{DA}, \; \textbf{\textrm{s}}^{DA}), \\
	s_t^{FR} = BiLSTM(e_t \oplus a_t^{DA}), \\
	a_t^{FR} = SelfAtt(s_t^{FR}, \; \textbf{\textrm{s}}^{FR}), \\ s_t^{AR} = BiLSTM(e_t \oplus a_t^{FR})
\end{gather*}
\noindent where $\oplus$ represents the vector concatenation operator, $e_t$ is the embedding of the word at time $t$, and $\textbf{\textrm{s}}^{L}$ = ($s_1^L$, ..., $s_T^L$) is the embedded sequence output of each $L$ layer, with $L = \{DA, FR, AR\}$. Given an input sentence, the final sequence of labels $\textbf{y}^L$ for each task is computed through a CRF tagging layer, which operates on the output of the $DA$ and $FR$ self-attention, and of the $AR$ BiLSTM embedding, so that:
\begin{gather*}
	\small
	\textbf{y}^{DA} = CRF^{DA}(\textbf{a}^{DA}), \; \textbf{y}^{FR} = CRF^{FR}(\textbf{a}^{FR}) \\
	\textbf{y}^{AR} = CRF^{AR}(\textbf{s}^{AR}),
\end{gather*}
where \textbf{a}$^{DA}$, \textbf{a}$^{FR}$ are attended embedded sequences.
Due to shortcut connections, layers in the upper levels of the architecture can rely both on direct word embeddings as well as the hidden representation $a_t^L$ computed by a previous layer. Operationally, the latter carries task specific information which, combined with the input embeddings, helps in stabilising the classification of each CRF layer, as shown by our experiments. The network is trained by minimising the sum of the individual negative log-likelihoods of the three CRF layers, while at test time the most likely sequence is obtained through the Viterbi decoding over the output scores of the CRF layer.

\section{Experimental Evaluation}
\label{sec:experiments}

In order to assess the effectiveness of the proposed architecture and compare against existing off-the-shelf tools, we run several empirical evaluations. 

\subsection{Datasets}\label{sec:data}
We tested the system on two datasets, different in size and complexity of the addressed language.
\paragraph{NLU-Benchmark dataset}
The first (publicly available) dataset, NLU-Benchmark (NLU-BM), contains $25,716$ utterances annotated with targeted \textit{Scenario}, \textit{Action}, and involved \textit{Entities}.
For example, ``\textit{schedule a call with Lisa on Monday morning}'' is labelled to contain a \texttt{calendar} scenario, where the \texttt{set\_event} action is instantiated through the entities [\texttt{event\_name}: \textit{a call with Lisa}] and [\texttt{date}: \textit{Monday morning}]. The Intent is then obtained by concatenating scenario and action labels (e.g., \texttt{calendar\_set\_event}).
This dataset consists of multiple home assistant task domains (e.g., scheduling, playing music), chit-chat, and commands to a robot \cite{Liu2019}.\footnote{Available at \url{https://github.com/xliuhw/NLU-Evaluation-Data}.}

\begin{table}[h]
	\centering
	\footnotesize
	\begin{tabular}{rcc}
		{} & {NLU-BM} & {NLU-BM (reduced)}\\
		\midrule
		\midrule
		{Sentences} & {$25715$} & {$11020$}\\
		{Sentences length} & {$7.06$} & {$6.84$}\\
		{Scenario labels set} & {$18$} & {$18$}\\
		{Action labels set} & {$54$} & {$51$}\\
		{Intent labels set} & {$68$} & {$64$}\\
		{Entity labels set} & {$56$} & {$54$}\\
		{Number of intent} & {$25715$} & {$11020$}\\
		{Number of entities} & {$20597$} & {$9130$}\\
		{Intents/sentence} & {$1$} & {$1$}\\
		{Entities/sentence} & {$0.8$} & {$0.83$}\\
	\end{tabular}
	\caption{Statistics of the NLU-Benchmark dataset \cite{Liu2019}.\label{tab:benchmark_statistics}}
\end{table}

\paragraph{\datasetname{} dataset}
The second dataset, \datasetname{}, is composed of $1,431$ sentences, for each of which dialogue acts, semantic frames, and corresponding frame elements are provided. 
This dataset is being developed for modelling user utterances to open-domain conversational systems for robotic platforms that are expected to handle different interaction situations/patterns -- e.g., chit-chat, command interpretation. The corpus is composed of different subsections, addressing heterogeneous linguistic phenomena, ranging from imperative instructions (e.g., ``\textit{enter the bedroom slowly, turn left and turn the lights off }'') to complex requests for information (e.g., ``\textit{good morning I want to buy a new mobile phone is there any shop nearby?}'') or open-domain chit-chat (e.g., ``\textit{nope thanks let's talk about cinema}'').
A considerable number of utterances in the dataset is collected through Human-Human Interaction studies in robotic domain ($\approx$$70\%$), though a small portion has been synthetically generated for balancing the frame distribution.

\begin{table}
	\centering
	\footnotesize
	\begin{tabular}{rc}
		{} & {\datasetname{} dataset}\\
		\midrule
		\midrule
		{Sentences} & {$1431$}\\
		{Sentences length} & {$7.24$}\\
		{Dialogue act labels set} & {$11$}\\
		{Frame labels set} & {$58$}\\
		{Frame element labels set} & {$84$}\\
		{Number of dialogue acts} & {$1906$}\\
		{Number of frames} & {$2013$}\\
		{Number of frame elements} & {$5059$}\\
		{Dialogue act/sentence} & {$1.33$}\\
		{Frames/sentence} & {$1.41$}\\
		{Frame elements/sentence} & {$3.54$}\\
	\end{tabular}
	\caption{Statistics of the \datasetname{} dataset.\label{tab:mummer_statistics}}
\end{table}

Note that while the NLU-BM is designed to have at most one intent per utterance, sentences are here tagged following the IOB2 sequence labelling scheme (see example of Figure \ref{fig:tagging}), so that multiple dialogue acts, frames, and frame elements can be defined at the same time for the same utterance. For example, three dialogue acts are identified within the sentence [\textit{good morning}]$_{\textsc{Opening}}$ [\textit{I want to buy a new mobile phone}]$_{\textsc{Inform}}$ [is there any shop nearby?]$_{\textsc{Req\_info}}$.
As a result, though smaller, the \datasetname{} dataset provides a richer representation of the sentence's semantics, making the tasks more complex and challenging. These observations are highlighted by the statistics in Table \ref{tab:mummer_statistics}, that show an average number of dialogue acts, frames and frame elements always greater than $1$ (i.e., $1.33$, $1.41$ and $3.54$, respectively).

\subsection{Experimental setup}
All the models are implemented with Keras~\cite{chollet2015keras} and Tensorflow~\cite{tensorflow2015-whitepaper} as backend, and run on a Titan Xp.
Experiments are performed in a 10-fold setting, using one fold for tuning and one for testing. However, since \architecturename{} is designed to operate on dialogue acts, semantic frames and frame elements, the best hyperparameters are obtained over the \datasetname{} dataset via a grid search using early stopping, and are applied also to the NLU-BM models.\footnote{Notice that in the NLU-BM experiments only the number of epochs is tuned, using $10\%$ of the training data.} This guarantees fairness towards other systems, that do not perform any fine-tuning on the training data.
We make use of pre-trained 1024-dim ELMo embeddings \cite{peters2018} as word vector representations without re-training the weights.

\subsection{Experiments on the NLU-Benchmark}
\label{sec:benchmark}
This section shows the results obtained on the NLU-Benchmark (NLU-BM) dataset provided by \cite{Liu2019}, by comparing \textbf{\architecturename{}} to off-the-shelf NLU services, namely: \textbf{Rasa}\footnote{\url{https://rasa.com/}}, \textbf{Dialogflow}\footnote{\url{https://dialogflow.com/}}, \textbf{LUIS}\footnote{\url{https://www.luis.ai/}} and \textbf{Watson}\footnote{\url{https://www.ibm.com/watson}}. In order to apply \architecturename{} to NLU-BM annotations, these have been aligned so that Scenarios are treated as DAs, Actions as FRs and Entities as ARs.

To make our model comparable against other approaches, we reproduced the same folds as in \cite{Liu2019}, where a resized version of the original dataset is used. Table \ref{tab:benchmark_statistics} shows some statistics of the NLU-BM and its reduced version. Moreover, micro-averaged Precision, Recall and F1 are computed following the original paper to assure consistency. TP, FP and FN of intent labels are obtained as in any other multi-class task. An entity is instead counted as TP if there is an overlap between the predicted and the gold span, and their labels match. 
\begin{table*}[ht]
	\centering
	\footnotesize
	\begin{tabular}{rccc|ccc}
		{} & \multicolumn{3}{c}{Intent} & \multicolumn{3}{c}{Entity}\\
		{} & P & R & F1 & P & R & F1\\
		\midrule
		\midrule
		{Rasa} & {$86.31 {\pm 1.07}$} & {$86.31 {\pm 1.07}$} & {$86.31 {\pm 1.07}$} & {$85.93 {\pm 1.05}$} & {$69.40 {\pm 1.66}$} & {$76.78 {\pm 1.27}$}\\
		{Dialogflow} & {$86.97 {\pm 2.02}$} & {$85.87 {\pm 2.33}$} & {$86.42 {\pm 2.18}$} & {$78.21 {\pm 3.35}$} & {$70.85 {\pm 4.70}$} & {$74.30 {\pm 3.74}$}\\
		{LUIS} & {$85.53 {\pm 1.14}$} & {$85.51 {\pm 1.15}$} & {$85.52 {\pm 1.15}$} & {$83.69 {\pm 1.31}$} & {$72.46 {\pm 2.05}$} & {$77.66 {\pm 1.45}$}\\
		{Watson\footnotemark} & {$\textbf{88.41} {\pm 0.68}$} & {$\textbf{88.08} {\pm 0.74}$} & {$\textbf{88.24} {\pm 0.70}$} & {$35.39 {\pm 0.93}$} & {$78.70 {\pm 2.01}$} & {$48.82 {\pm 1.14}$}\\
		\midrule
		{\textbf{\architecturename{}}} & {$87.41 {\pm 0.63}$} & {$87.70 {\pm 0.64}$} & {$87.55 {\pm 0.63}$} & {$\textbf{87.65} {\pm 0.98}$} & {$\textbf{82.04} {\pm 2.12}$} & {$\textbf{84.74} {\pm 1.18}$}\\
	\end{tabular}
	\caption{Comparison of \architecturename{} with the results obtained in~\cite{Liu2019} for Intents and Entity Types.\label{tab:benchmark}}
\end{table*}

Experimental results are reported in Table \ref{tab:benchmark}. The statistical significance is evaluated through the Wilcoxon signed-rank test.
When looking at the intent F1, \architecturename{} performs significantly better than Rasa $[Z=-2.701, p = .007]$ and LUIS $[Z=-2.807, p = .005]$.
On the contrary, the improvements w.r.t. Dialogflow $[Z=-1.173, p = .241]$ do not seem to be significant. This is probably due to the high variance obtained by Dialogflow across the 10 folds.
Watson is by a significant margin the most accurate system in recognising intents $[Z=-2.191, p = .028]$, especially due to its Precision score.

The hierarchical multi-task architecture of \architecturename{} seems to contribute strongly to entity tagging accuracy. In fact, in this task it performs significantly better than Rasa $[Z=-2.803, p = .005]$, Dialogflow $[Z=-2.803, p = .005]$, LUIS $[Z=-2.803, p = .005]$ and Watson $[Z=-2.805, p = .005]$, with improvements from $7.08$ to $35.92$ of F1.\footnote{Results for Watson are shown for the non-contextual training. Due to Watson limitations, i.e. 2000 training examples for contextual training, we could not run the whole test in such configuration. For fairness, we report results made on 8 random samplings of 2000/1000 train/test examples a each (F1): Intent=$72.64\pm7.46$, Slots=$77.01\pm10.65$, Combined=$74.85\pm7.54$}

\begin{table}[ht]
	\centering
	\footnotesize
	\resizebox{\columnwidth}{!}{%
		\begin{tabular}{lccc}
			{} & \multicolumn{3}{c}{Combined}\\
			{} & P & R & F1\\
			\midrule
			\midrule
			{Rasa} & {$86.16 {\pm 0.90}$} & {$78.66 {\pm 1.28}$} & {$82.24 {\pm 1.08}$}\\
			{Dialogflow} & {$83.19 {\pm 2.43}$} & {$79.07 {\pm 3.10}$} & {$81.07 {\pm 2.64}$}\\
			{LUIS} & {$84.76 {\pm 0.67}$} & {$79.61 {\pm 1.25}$} & {$82.1 {\pm 0.90}$}\\
			{Watson} & {$54.02 {\pm 0.75}$} & {$83.83 {\pm 1.02}$} & {$65.7 {\pm 0.75}$}\\
			\midrule
			{\textbf{\architecturename{}}} & {$\textbf{87.52} {\pm 0.61}$} & {$\textbf{85.03} {\pm 1.11}$} & {$\textbf{86.25} {\pm 0.66}$}\\
		\end{tabular}
	}
	\caption{Comparison of \architecturename{} with the results in~\cite{Liu2019} by combining Intent and Entity.\label{tab:benchmark_combined}}
\end{table}

Following \cite{Liu2019}, we then evaluated a metric that combines intent and entities, computed by simply summing up the two confusion matrices (Table \ref{tab:benchmark_combined}).
Results highlight the contribution of the entity tagging task, where \architecturename{} outperforms the other approaches. Paired-samples t-tests were conducted to compare the \architecturename{} combined F1 against the other systems. The statistical analysis shows a significant improvement over Rasa $[Z=-2.803, p = .005]$, Dialogflow $[Z=-2.803, p = .005]$, LUIS $[Z=-2.803, p = .005]$ and Watson $[Z=-2.803, p = .005]$.

\subsubsection{Ablation study}
In order to assess the contributions of the \architecturename's components, we performed an ablation study. The results are obtained on the NLU-BM, following the same setup as in Section \ref{sec:benchmark}.

\begin{table}[ht]
	\centering
	\footnotesize
	\resizebox{\columnwidth}{!}{%
		\begin{tabular}{lccc}
			{} & {Intent} & {Entity} & \parbox{1.35cm}{Combined}\\
			\midrule
			\midrule
			\parbox{2cm}{\textbf{HERMIT}} & \parbox{1.35cm}{$\textbf{87.55} {\pm 0.63}$} & \parbox{1.35cm}{$84.74 {\pm 1.18}$} & \parbox{1.35cm}{$\textbf{86.25} {\pm 0.66}$}\\
			\midrule
			\parbox{2cm}{\enskip-- SA} & \parbox{1.35cm}{$87.03 {\pm 0.74}$} & \parbox{1.35cm}{$84.35 {\pm 1.15}$} & \parbox{1.35cm}{$85.81 {\pm 0.81}$}\\
			\parbox{2cm}{\enskip-- SA/CN} & \parbox{1.35cm}{$87.09 {\pm 0.78}$} & \parbox{1.35cm}{$82.43 {\pm 1.42}$} & \parbox{1.35cm}{$84.97 {\pm 0.72}$}\\
			\parbox{2cm}{\enskip-- SA/CRF} & \parbox{1.35cm}{$83.57 {\pm 0.75}$} & \parbox{1.35cm}{$\textbf{84.77} {\pm 1.06}$} & \parbox{1.35cm}{$84.09 {\pm 0.79}$}\\
			\parbox{2cm}{\enskip-- SA/CN/CRF} & \parbox{1.35cm}{$83.78 {\pm 1.10}$} & \parbox{1.35cm}{$82.22 {\pm 1.41}$} & \parbox{1.35cm}{$83.10 {\pm 1.06}$}\\
		\end{tabular}
	}
	\caption{Ablation study of \architecturename{} on the NLU-BM.\label{tab:benchmark_ablation}}
\end{table}

Results are shown in Table \ref{tab:benchmark_ablation}. The first row refers to the complete architecture, while {--SA} shows the results of \architecturename{} without the self-attention mechanism. Then, from this latter we further remove shortcut connections ({-- SA/CN}) and CRF taggers ({-- SA/CRF}). The last row ({-- SA/CN/CRF}) shows the results of a simple architecture, without self-attention, shortcuts, and CRF. Though not significant, the contribution of the several architectural components can be observed.
The contribution of self-attention is distributed across all the tasks, with a small inclination towards the upstream ones.
This means that while the entity tagging task is mostly lexicon independent, it is easier to identify pivoting keywords for predicting the intent, e.g. the verb \textit{``schedule''} triggering the \texttt{calendar\_set\_event} intent.
The impact of shortcut connections is more evident on entity tagging. In fact, the effect provided by shortcut connections is that the information flowing throughout the hierarchical architecture allows higher layers to encode richer representations (i.e., original word embeddings + latent semantics from the previous task).
Conversely, the presence of the CRF tagger affects mainly the lower levels of the hierarchical architecture. This is not probably due to their position in the hierarchy, but to the way the tasks have been designed. In fact, while the span of an entity is expected to cover few tokens, in intent recognition (i.e., a combination of Scenario and Action recognition) the span always covers all the tokens of an utterance. CRF therefore preserves consistency of IOB2 sequences structure.
However, \architecturename{} seems to be the most stable architecture, both in terms of standard deviation and task performance, with a good balance between intent and entity recognition.

\subsection{Experiments on the \datasetname{} dataset}
\begin{table*}
	\centering
	\footnotesize
	\begin{tabular}{rccc|cc}
		{} & P & R & F1 & span F1 & EM\\
		\midrule
		\midrule
		{\textit{Dialogue act}} & {$96.49 {\pm 0.98}$} & {$95.95 {\pm 1.41}$} & {$96.21 {\pm 1.13}$} & {$89.42 {\pm 3.74}$} & {$89.31 {\pm 3.28}$}\\
		{\textit{Frame}} & {$95.26 {\pm 0.95}$} & {$94.02 {\pm 1.20}$} & {$94.64 {\pm 1.09}$} & {$84.40 {\pm 2.99}$} & {$82.60 {\pm 2.68}$}\\
		{\textit{Frame element}} & {$95.62 {\pm 0.61}$} & {$93.98 {\pm 0.76}$} & {$94.79 {\pm 0.56}$} & {$92.26 {\pm 1.22}$} & {$79.73 {\pm 2.03}$}\\
		{\textbf{Combined}} & {$93.90 {\pm 0.89}$} & {$92.95 {\pm 0.86}$} & {$93.42 {\pm 0.83}$} & {--} & {$\textbf{69.53} {\pm 2.50}$}\\
	\end{tabular}
	\caption{\architecturename{} performance over the \datasetname{} dataset. P,R and F1 are evaluated following \cite{Liu2019} metrics\label{tab:mummer}}
\end{table*}

In this section we report the experiments performed on the \datasetname{} dataset (Table \ref{tab:mummer}). Together with the evaluation metrics used in \cite{Liu2019}, we report the span F1, computed using the CoNLL-2000 shared task evaluation script, and the Exact Match (EM) accuracy of the entire sequence of labels. It is worth noticing that the EM Combined score is computed as the conjunction of the three individual predictions -- e.g., a match is when all the three sequences are correct.

Results in terms of EM reflect the complexity of the different tasks, motivating their position within the hierarchy. Specifically, dialogue act identification is the easiest task ($89.31\%$) with respect to frame ($82.60\%$) and frame element ($79.73\%$), due to the shallow semantics it aims to catch. However, when looking at the span F1, its score ($89.42\%$) is lower than the frame element identification task ($92.26\%$). What happens is that even though the label set is smaller, dialogue act spans are supposed to be longer than frame element ones, sometimes covering the whole sentence. Frame elements, instead, are often one or two tokens long, that contribute in increasing span based metrics.
Frame identification is the most complex task for several reasons. First, lots of frame spans are interlaced or even nested; this contributes to increasing the network entropy. Second, while the dialogue act label is highly related to syntactic structures, frame identification is often subject to the inherent ambiguity of language (e.g., \textit{get} can evoke both \textit{Commerce\_buy} and \textit{Arriving}).
We also report the metrics in \cite{Liu2019} for consistency. For dialogue act and frame tasks, scores provide just the extent to which the network is able to detect those labels. In fact, the metrics do not consider any span information, essential to solve and evaluate our tasks. However, the frame element scores are comparable to the benchmark, since the task is very similar.

Overall, getting back to the combined EM accuracy, \architecturename{} seems to be promising, with the network being able to reproduce all the three gold sequences for almost $70\%$ of the cases. The importance of this result provides an idea of the architecture behaviour over the entire pipeline.

\subsection{Discussion}
\label{sec:discussion}
The experimental evaluation reported in this section provides different insights. The proposed architecture addresses the problem of NLU in wide-coverage conversational systems, modelling semantics through multiple Dialogue Acts and Frame-like structures in an end-to-end fashion. In addition, its hierarchical structure, which reflects the complexity of the single tasks, allows providing rich representations across the whole network. In this respect, we can affirm that the architecture successfully tackles the multi-task problem, with results that are promising in terms of usability and applicability of the system in real scenarios.

However, a thorough evaluation in the wild must be carried out, to assess to what extent the system is able to handle complex spoken language phenomena, such as repetitions, disfluencies, etc. To this end, a real scenario evaluation may open new research directions, by addressing new tasks to be included in the multi-task architecture. This is supported by the scalable nature of the proposed approach. Moreover, following \cite{sanh2018hmtl}, corpora providing different annotations can be exploited within the same multi-task network.

We also empirically showed how the same architectural design could be applied to a dataset addressing similar problems. In fact, a comparison with off-the-shelf tools shows the benefits provided by the hierarchical structure, with better overall performance better than any current solution. An ablation study has been performed, assessing the contribution provided by the different components of the network. The results show how the shortcut connections help in the more fine-grained tasks, successfully encoding richer representations. CRFs help when longer spans are being predicted, more present in the upstream tasks.

Finally, the seq2seq design allowed obtaining a multi-label approach, enabling   the identification of multiple spans in the same utterance that might evoke different dialogue acts/frames. This represents a novelty for NLU in conversational systems, as such a problem has always been tackled as a single-intent detection. However, the seq2seq approach carries also some limitations, especially on the Frame Semantics side. In fact, label sequences are linear structures, not suitable for representing nested predicates, a tough and common problem in Natural Language. For example, in the sentence ``\textit{I want to buy a new mobile phone}'', the [\textit{to buy a new mobile phone}] span represents both the \textsc{Desired\_event} frame element of the \textit{Desiring} frame and a \textit{Commerce\_buy} frame at the same time. At the moment of writing, we are working on modeling nested predicates through the application of bilinear models.

\section{Future Work}

We have started integrating a corpus of 5M sentences of real users chit-chatting with our conversational agent, though at the time of writing they represent only $16\%$ of the current dataset.

As already pointed out in Section \ref{sec:discussion}, there are some limitations in the current approach that need to be addressed.
First, we have to assess the network's capability in handling typical phenomena of spontaneous spoken language input, such as repetitions and disfluencies \cite{Shalyminov.etal18}. This may open new research directions, by including new tasks to identify/remove any kind of noise from the spoken input.
Second, the seq2seq scheme does not deal with nested predicates, a common aspect of Natural Language. To the best of our knowledge, there is no architecture that implements an end-to-end network for FrameNet based semantic parsing. Following previous work \cite{Strubel18:emnlp}, one of our future goals is to tackle such problems through hierarchical multi-task architectures that rely on bilinear models.

\section{Conclusion}
In this paper we presented \architecturename{} NLU, a hierarchical multi-task architecture for semantic parsing sentences for cross-domain spoken dialogue systems. The problem is addressed using a seq2seq model employing BiLSTM encoders and self-attention mechanisms and followed by CRF tagging layers. We 
evaluated \architecturename{} on a 25K sentences NLU-Benchmark and out-perform state-of-the-art NLU tools such as Rasa, Dialogflow, LUIS and Watson, even without specific fine-tuning of the model.

\section*{Acknowledgement}
This research was partially supported by the European Union's Horizon 2020 research and innovation programme under grant agreement No.\ 688147 (MuMMER project\footnote{\url{http://mummer-project.eu/}}).

\bibliography{main}
\bibliographystyle{acl_natbib}

\end{document}